\documentclass[10pt, a4paper]{article}
\usepackage{lrec}
\usepackage{multibib}
\newcites{languageresource}{Language Resources}
\usepackage{graphicx}
\usepackage{tabularx}
\usepackage{soul}

\usepackage{epstopdf}
\usepackage[utf8]{inputenc}
\usepackage[T1]{fontenc}
\usepackage[italian,polish,czech,english]{babel}

\usepackage{hyperref}
\usepackage{xstring}
\usepackage{booktabs}
\usepackage{color}
\usepackage{xurl}
\usepackage[dvipsnames]{xcolor}
\usepackage{footnote}

\title{O-Dang! The Ontology of Dangerous Speech Messages}

\name{Marco A. Stranisci$^{\ast}$, Simona Frenda$^{\ast\diamond}$, Mirko Lai$^{\ast}$, Oscar Araque$^{\star}$, Alessandra T. Cignarella$^{\ast}$,\\ 
\bf \large Valerio Basile$^{\ast}$, Viviana Patti$^{\ast}$, Cristina Bosco$^{\ast}$} 

\address{$^{\ast}$ Department of Computer Science - University of Turin, Turin, Italy \\
$^{\diamond}$ PRHLT Research Center - Universitat Polit\`ecnica de Val\`encia, Valencia, Spain\\
$^{\star}$ Intelligent Systems Group - ETSI Telecomunicación - Universidad Politécnica de Madrid, Madrid, Spain\\
{\tt \small \{marcoantonio.stranisci simona.frenda, mirko.lai, alessandrateresa.cignarella,}\\
{\tt \small valerio.basile, viviana.patti, cristina.bosco\}@unito.it}\\
{\tt \small o.araque@upm.es}}

\abstract{
Inside the NLP community there is a considerable amount of language resources created, annotated and released every day with the aim of studying specific linguistic phenomena. Despite a variety of attempts in order to organize such resources has been carried on, a lack of systematic methods and of possible interoperability between resources are still present. Furthermore, when storing linguistic information, still nowadays, the most common practice is the concept of ``gold standard'', which is in contrast with recent trends in NLP that aim at stressing the importance of different \textit{subjectivities} and points of view when training machine learning and deep learning methods.
In this paper we present O-Dang!: The Ontology of Dangerous Speech Messages, a systematic and interoperable Knowledge Graph (KG) for the collection of linguistic annotated data. 
O-Dang! is designed to gather and organize 
Italian datasets 
into a structured KG, according to the principles shared within the Linguistic Linked Open Data community. The ontology has also been designed to account
a \textit{perspectivist} approach, since it provides a model for encoding 
both gold standard and single-annotator labels in the KG.
The paper is structured as follows. In Section \ref{sec:intro} the motivations of our work are outlined. Section \ref{sec:onto} describes the O-Dang! Ontology, that provides a common semantic model for the integration of datasets in the KG. The Ontology Population stage with information about corpora, users, and annotations is presented in Section \ref{sec:kgpopulation}. Finally, in Section \ref{sec:hurtlex} an analysis of offensiveness across corpora is provided as a first case study for the resource.
\\ \newline 
}

\begin{document}
\maketitleabstract



    






\section{Introduction and Motivation}\label{sec:intro}
In this day and age, in almost every research field -- as well as in Computational Linguistics -- it is considered an enormous wealth to have the presence of manually annotated data sets in order to implement Machine Learning and Deep Learning pipelines. In the last 15 years there has been a very extensive effort within many research groups that deal with 
Natural Language Processing (NLP) for the creation, development and maintenance of corpora of linguistic data annotated with regard to various phenomena.

Nowadays, there are thousands of data sets that model similar phenomena in many different languages, and it often happens that each research group models a phenomenon on the basis of their own annotation scheme, usually not shared with other researchers, who are involved in studying similar phenomena on different languages. Another frequent case in the modeling of linguistic phenomena is 
to develop new annotations 
adding further layers of information on top of pre-existing ones to train, for instance, models based on multitask learning.

The research idea we would like to present in this paper stems from the need to provide a more structured organization to the myriad of linguistic resources and datasets developed in the NLP field, and to guarantee interoperability and dialogue between similar resources. 

Among the many projects that already devoted their efforts in creating a bridge between sentiment and emotion analysis and linguistic data, we mostly referred to EUROSENTIMENT \cite{sanchez2014eurosentiment}
developing a common language resource representation model based on established Linked Data formats such as Onyx \citelanguageresource{sanchez2016onyx} and Marl \citelanguageresource{westerki2013marl}. 

In this paper, we describe the creation of a Linguistic Linked Open Data (LLOD) resource, focused on collecting dangerous messages that indirectly contribute to the spread of discriminatory contents, thus called \textbf{The Ontology of Dangerous Speech} (O-Dang!).

\emph{Dangerous Speech} has been defined by \newcite{benesch2012dangerous} as a speech that ``has a reasonable \textit{chance} of catalyzing or amplifying violence by one group against another, given the circumstances in which it was made or disseminated''. 
This \textit{chance} materializes when the circumstances in which the speech takes place consist of: 1) a powerful speaker or source with a high degree of influence, 2) an audience that believe to be subject to a threat, 3) a social and historical context propitious for the violence, 4) the means of dissemination (such as social media), 5) the content of the speech that aims at the process of dehumanization, guilt attribution, threat construction, destruction of alternatives, creation of a new semantics of the violence conceived as admirable, linked to praiseworthy qualities and based on specific biased references that justify it \cite{leader2016dangerous}. Dangerous speech, therefore, is a type of speech that aims at contributing to create a climate of violence and intolerance against protected groups of people, such as women, immigrants, religious minorities, and others.

\begin{figure*}
    \centering
    \includegraphics[width=.7\textwidth]{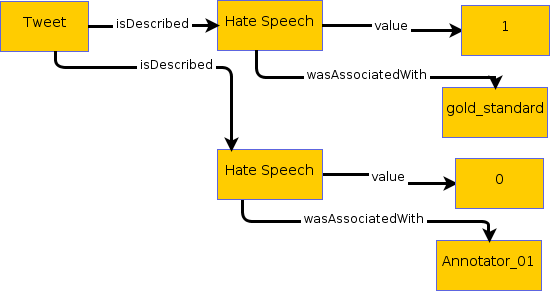}
    \caption{A snapshot of the O-Dang Semantic Model in which gold standard and un-aggregated annotations are encoded.}
    \label{fig:tweet}
\end{figure*}

As some scholars highlighted, there are various rhetorical and pragmatic devices that play a part in the expression of dangerous utterances. For instance, \newcite{grimminger2021hate} and \newcite{frenda2019stance} reflected on the use of offensive and toxic communication in tweets expressing a stance towards specific political candidates (such as Biden and Trump) or sensible social issues involving a particular target such as women (like feminist movements or abortion). Others focused more on the use of the ironic language to lessen the negative tones of the hateful messages, making their automatic recognition challenging \cite{nobata2016abusive,FRENDA2022116398}. The employment of these kinds of devices actually lets speakers or users to be less explicit in their claims, limiting, thus, their exposure.

From this perspective, 
we designed an ontology for storing existing 
Italian corpora in a Knowledge Graph (KG) that is interoperable and that takes into account 
general characteristics of the various NLP datasets annotated for various dimensions of hate such as Hate Speech (HS), misogyny, stereotypes, and offensive and aggressive language. 
The semantic model is general enough to 
populate the KG with other corpora focused on orthogonal phenomena to hate, such as stance or ironic language, realizing a tool that is open to collaborative effort of the scientific community.

In this sense we are inspired by the work of \newcite{bender2018datastatement} in which the authors propose data statements as a design solution and professional practice for natural language processing technologists to be followed when creating a linguistic resource and making it available to other researchers.

Furthermore, our work follows the directives of the \textit{Perspectivist Data Manifesto}\footnote{\url{https://pdai.info/}.} \cite{basile2020s}, and that is, we do not limit ourselves to consider the data of written texts and the gold standard labels, but -- where possible -- we try to store in the KG the labels of the different annotations in un-aggregated form for emphasizing the importance of the different perspectives and points of view of individual subjectivities of human annotators. 
Finally, our work is also inspired by \newcite{lewandoska2021ontologyoffensive} which is focused on aligning several phenomena correlated to discrimination in a unique semantic model. 

The contributions to be found in this article are the following:

\begin{itemize}
\item \textbf{The O-Dang! Ontology}\footnote{\url{https://github.com/marcostranisci/o-dang}}, a semantic model aimed at describing and linking a variety of datasets containing Dangerous Speech and orthogonal phenomena;

\item A KG containing $11$ existing Italian NLP data sets on Dangerous Speech and parallel phenomena. The KG serves as a first case study for providing interoperability between corpora annotatet fo Dangerous Speech;

\item  \textbf{an Entity Linking pipeline} for recovering the specific targets of Dangerous Speech and abusive language;

\item \textbf{un-aggregated annotations} of the datasets developed by our research group in the past years;


\end{itemize}

The resulting KG will be available through endpoint SPARQL, allowing several applications, among which:
\begin{itemize}

\item exporting personalized portion of the KG for the study of specific phenomena across corpora and the configuration of different training sets. 

\item querying all Dangerous Speech referred to specific persons and groups

\item filtering gold standard and un-aggregated annotation

\item querying the communication interaction among users and messages
\end{itemize}

\begin{figure*}
    \centering
    \includegraphics[width=.7\textwidth]{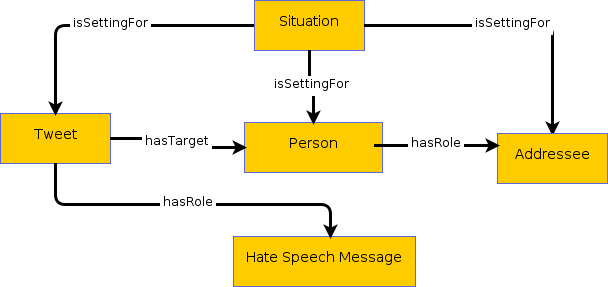}
    \caption{A portion of the O-Dang Semantic Model where a communicative situation with a participant is represented.}
    \label{fig:situation}
\end{figure*}

\begin{savenotes}
\begin{table*}[ht]
    \centering
    \resizebox{\textwidth}{!}{
    \begin{tabular}{lllllr}
    \textbf{name} & \textbf{reference} & \textbf{genre} & \textbf{phenomena} & \textbf{annotation} & \textbf{size} \\
    \toprule
    IronITA 2018 & 
    \citelanguageresource{cignarella2018overview} & 
    tweets & 
    \begin{tabular}[c]{@{}l@{}}
    irony (0/1), \\ 
    sarcasm (0/1),\\
    --- \\
    for some data: \\ 
    type of irony\footnote{Implicit or explicit.} \\
    category of irony\footnote{Analogy, euphemism, context shift, false assertion, hyperbole, oxymoron-paradox, rhetorical question, other.}\\
    PoS tags \& UD \end{tabular} & 
    un-aggregated & 
    $4,849$ \\
    \midrule
    AMI 2018 & 
    \citelanguageresource{fersini2018overview} & 
    tweets & 
    \begin{tabular}[c]{@{}l@{}}
    misogyny (0/1), \\ 
    category (stereotype / dominance \\
    derailing / sexual harassment \\
    discredit), \\ 
    target classification (active/passive) \end{tabular} & 
    aggregated & 
    $5,000$ \\
    \midrule
    HaSpeeDe 2018 & 
    \citelanguageresource{sanguinetti2018italian} &
    \begin{tabular}[c]{@{}l@{}}
    facebook posts \\ 
    and tweets \end{tabular} & 
    hate speech (0/1) & 
    aggregated  & 
    $7,996$ \\
    \midrule
    Hate Speech Corpus & 
    \citelanguageresource{sanguinetti2018italian} & 
    tweets & 
    \begin{tabular}[c]{@{}l@{}}
    hate speech (0/1), \\ 
    stereotype (0/1), \\ 
    aggressiveness (0/1), \\ 
    irony (0/1), \\ 
    intensity (0--\textgreater{}4) \end{tabular} & 
    un-aggregated & 
    $6,928$ \\
    \midrule
    SardiStance 2020 & \citelanguageresource{cignarella2020sardistance} & 
    tweets & 
    \begin{tabular}[c]{@{}l@{}}
    stance (against/favor/none) \\ 
    irony (0/1) \end{tabular} & 
    un-aggregated & 
    $3,242$ \\
    \midrule
    AMI 2020 & 
    \citelanguageresource{fersini2020ami} & 
    tweets & 
    \begin{tabular}[c]{@{}l@{}}
    misogyny (0/1) \\
    aggressiveness (0/1) \end{tabular} & 
    aggregated & 
    $7,000$ \\
    \midrule
    HaSpeeDe 2020 & 
    \citelanguageresource{sanguinetti2020haspeede} & 
    tweets and news headlines & 
    \begin{tabular}[c]{@{}l@{}}
    hate speech (0/1), \\ 
    stereotype (0/1), \\ 
    aggressiveness (0/1), \\ 
    irony (0/1), \\ 
    sarcasm (0/1)\\ 
    --- \\
    for some data: \\ 
    offensiveness (0/1), \\ 
    intensity (0--\textgreater{}4) \\ 
    nominal utterances \end{tabular} & 
    un-aggregated &
    $8,602$       \\
    \midrule
    Moral ConvITA & \citelanguageresource{stranisci2021expression} & tweets & moral stance & un-aggregated & $1,722$ \\
    \midrule
    Populismo Penale & N/A & tweets & stance (against/favor/none) & un-aggregated & $12,479$ \\
    \midrule
    \\
    Silvia Romano Corpus & 
    N/A & 
    tweets & 
    stance, abusive language & 
    un-aggregated & 
    $4,913$ \\
    \midrule
    Crowd-HS & N/A & tweets & hate speech (0-7) & un-aggregated & $926$ \\
    \bottomrule
    \end{tabular}
    } 
    
    \caption{Summary of the datasets which are already populating O-Dang!}
    \label{tab:dataset_inside_ontology}
\end{table*}
\end{savenotes}

\section{The O-Dang! Ontology} \label{sec:onto}
The O-Dang! Ontology provides a general encoding for the harmonization of different datasets in a unique resource. The model relies on existing authoritative resources, such as Dolce \cite{gangemi2002sweetening}, Prov-O \cite{lebo2013prov}, and FRBR \cite{tillett2005frbr}\footnote{Prefixes of existing ontologies reused in our model are the following: Dolce (dul), Prov-O (prov), FRBR (frbr). Properties of classes of O-Dang! are introduced by `:'}, and represents three aspects about data: (i) the encoding of the annotated text, (ii) the provenance of annotations, (iii) the conversational situation in which the annotated message is present. 

A message is encoded as a \textsc{frbr:Expression} embedded in one or more \textsc{frbr:Manifestation} and linked to one or more annotated corpora through the property \textbf{dul:isPartOf}. 

All annotation schemes are represented as subclasses of \textsc{dul:Description}, since each scheme may be intended as a shared description of a concept between researchers and annotators. As it can be observed in Figure \ref{fig:tweet}, a \textsc{dul:isDescribed} as `Hate Speech' with a specific value. Such a modeling enables the comparison of different schemes adopted for annotating same concepts \cite{poletto2021resources} (eg: binary, scalar). Finally, the \textbf{prov:wasAssociatedWith} property links all annotations to their annotators 
Figure \ref{fig:tweet} shows two types of annotator: gold\_standard, namely a label to identify all aggregated annotations, and a set of individual annotators for researchers interested in querying un-aggregated data from the KG. It is important to notice that no socio-demographic information about annotators is provided within O-Dang!, but only anonymized ids of the type `annotator\_\textit{n}'.

Messages annotated as expressing a given concept are also encoded within a conversational situation, that is a \textbf{dul:Situation} in which people, messages, and groups may hold a role. Such representation is focused on the interaction between messages, concept related to Dangerous Speech, and Agents, allowing to query all messages that express a given phenomenon, and have a specific category as a target. In Figure \ref{fig:situation} the representation of an HS message may be observed. The Situation \textsc{dul:isSettingFor} a Tweet with the role of Hate Speech Message, which is the result of the annotation process depicted in \ref{fig:tweet}. The target of this message has also setting in the same situation with the role of Addressee. Below, an example of materialized triples encoding a HS message against Cécile Kyenge\footnote{She is an Italian politician and ex member of the European Parliament.} is provided.

\begin{verbatim} 
odang_situation_1342 a :Situation; 
    isSettingFor :@ckyenge;
    :isSettingFor [
    a :Tweet;
    :hasRole HateSpeechMessage;
    :hasText ``@ckyenge per fare 
    sentire     a casa voi africani
    e musulmani e stranieri'';
    :hasTarget :@ckyenge
    ];
    @ckyenge a :Person;
    :hasRole :Addressee;
    :gender :female;
    :citizenship :ITA;
    :placeOfBirth :Kambove.
\end{verbatim}


\section{Datasets and Ontology Population} \label{sec:kgpopulation}
The O-Dang! KG includes $898,016$ triples about $62,193$ tweets and $21,972$ users. The Ontology Population stage was performed in two steps: the integration of different data sets in the KG, and a Entity Linking pipeline for the population of the ontology with socio-demographic information about users who are target of Dangerous Speech.
\subsection{Dataset Integration}
Table \ref{tab:dataset_inside_ontology} shows the datasets that are already populating O-Dang!. 
As said in Section~\ref{sec:intro}, these corpora are related to Dangerous Speech and parallel phenomena such as irony and stance.
For each dataset, we provide the bibliographic reference, textual genres of data, the considered phenomena and the values used to label their presence, and finally the type of annotation (`aggregated' and `un-aggregated') provided by authors.
The un-aggregated annotations reveal the different perspectives or subjectivities on the perception of Dangerous Speech, as well as the difficulty of annotation of the phenomenon and, consequently, of ambiguous cases.
For instance, the following news headline (Example 1) was annotated by \texttt{annotator\_1} as hateful and by \texttt{annotator\_2} as non-hateful.

\begin{itemize}
    \item[(1)] \textit{Alessandria, straniero con ascia e martello aggredisce coppia in casa}\\
    $\rightarrow$Alexandria, a foreigner with ax and hammer attacks a couple at home
\end{itemize}

Beyond the clearness of the guidelines, the interpretation of the instances is subjective and relies on the backgrounds of the annotators \cite{Sohail_2021}.

\subsection{Entity Linking pipeline}
Information about addressees of dangerous messages from Twitter is provided in the KG through an Entity Linking pipeline. Names of each user who is mentioned in a reply have been retrieved through the Twitter API and then searched using Google KG. After a disambiguation process relying on exact string matching between the name provided in input and Google KG output, and on the Google KG score, all the corresponding Wikidata ID were retrieved. Finally, sociodemographic information about each user has been collected from Wikidata. The resulting number of users mapped within the KG is $344$. For each, the following information are provided: date of birth, place of birth, country of citizenship, sex or gender, occupation, political party. Below, an example of user associated with such properties is shown.

\begin{verbatim} 
odang_usr_7986 a :Person; 
    :hasID 322933929;
    :gender female;
    :birthYear 1985;
    :countryOfCitizenship :ITA;
    :placeOfBirth :Lugano;
    :occupation :politician;
    :politicalParty :DemocraticParty .
\end{verbatim}

\section{Lexical Analysis} \label{sec:hurtlex}

To perform lexical analysis catching the offensiveness of the messages contained in the datasets that at the moment populate O-Dang!, we employed HurtLex \citelanguageresource{bassignana2018hurtlex}.  
HurtLex  is a multilingual lexicon of hateful words created from the Italian lexicon ``Le Parole per Ferire'' by Tullio de Mauro. The entries in the lexicon are categorized in 17 types of offenses (see Table~\ref{tab:hurtlex-categories}) enclosed in two macro-categories: \textit{conservative} (words with literally offensive sense) and \textit{inclusive} (words with not literally offensive sense, but that could be used with negative connotation).
In particular, we considered only the conservative version of the hurtful categories which have been mapped within O-Dang! through OntoLex-Lemon \cite{mccrae2017ontolex}. Each \textit{conservative} word in HurtLex is represented as the following:

\begin{verbatim}
    :IT1241 a :LexicalEntry;
    rdfs:label `fannullone'/`loafer';
    lexinfo:partOfSpeech :Noun;
    :isDescribed :dmc.
    :dmc a :Offensive;
    rdfs:label `moral defects' .
\end{verbatim}

The idea is to exploit
HurtLex as a means to cross-evaluate the offensiveness of the datasets in the KG (even those which are not annotated expressly as dangerous), and to provide a further description of them.

\begin{table}[htb]
\centering
\begin{tabular}{lcp{4.8cm}}
   \textbf{category} & \textbf{length} & \textbf{description} \\
\hline
PS & 254 & Ethnic Slurs \\
RCI& 36 & Location and Demonyms \\
PA&167 & Profession and Occupation \\
DDP& 496 & Physical Disabilities and Diversity \\
DDF & 80& Cognitive Disabilities and Diversity \\
DMC& 657 & Moral Behavior and Defect \\
IS& 161 & Words Related to Social and Economic advantages\\
OR & 144& Words Related to Plants \\
AN & 775 & Words Related to Animals \\
ASM &303 & Words Related to Male Genitalia \\
ASF& 191 & Words Related to Female Genitalia \\
PR & 138& Words Related to Prostitution \\
OM &145& Words Related to Homosexuality \\
QAS&536 & Descriptive Words with Potential Negative Connotations \\
CDS&2042 & Derogatory Words \\
RE&391 & Felonies and Words Related to Crime and Immoral Behavior \\
SVP& 424 & Words Related to the Seven Deadly Sins of the Christian Tradition \\
 \hline
\end{tabular}    
\caption{HurtLex Categories.}
\label{tab:hurtlex-categories}
\end{table}

\begin{table*}[ht!]
\centering
\begin{tabular}{l||ccccccccc}
   \textbf{Dataset} & \textbf{PS} & \textbf{DDP} & \textbf{DDF} & \textbf{DMC} & \textbf{ASM} & \textbf{ASF} & \textbf{QAS} & \textbf{CDS} & \textbf{SVP} \\
\hline
IronITA 2018         &  0.0080 &  0.0217 &  0.0031 &  0.0431 &  0.0151 &  0.0103 &  0.0210 &  0.0693 &  0.0054 \\
HaSpeeDe 2018         &  0.0228 &  0.0430 &  0.0015 &  0.0540 &  0.0313 &  0.0180 &  0.0350 &  0.1286 &  0.0088 \\
Hate Speech Corpus                  &  0.0172 &  0.0345 &  0.0013 &  0.0453 &  0.0240 &  0.0141 &  0.0282 &  0.1068 &  0.0079 \\
SardiStance 2020  &  \textbf{0.0275} &  0.0584 &  0.0050 &  \textbf{0.0617} &  0.0413 &  0.0333 &  0.0333 &  0.1275 &  0.0110 \\
HaSpeeDe 2020         &  0.0199 &  0.0396 &  0.0016 &  0.0548 &  0.0263 &  0.0149 &  0.0307 &  0.1128 &  0.0074 \\
Moral ConvITA &  0.0175 &  0.0420 &  0.0014 &  0.0467 &  0.0236 &  0.0175 &  0.0331 &  0.1176 &  0.0080 \\
Populismo Penale     &  0.0142 &  0.0454 &  0.0015 &  0.0402 &  0.0156 &  0.0260 &  0.0226 &  \textbf{0.1478} &  0.0064 \\
Silvia Romano Corpus &  0.0196 &  \textbf{0.0589} &  0.0036 &  0.0429 &  \textbf{0.0429} &  \textbf{0.0339} &  \textbf{0.0405} &  0.1024 &  \textbf{0.0107} \\
Crowd-HS             &  0.0151 &  0.0356 &  0.0011 &  0.0508 &  0.0140 &  0.0130 &  0.0313 &  0.1274 &  0.0097 \\
\hline
\end{tabular}    
\caption{Characterization of the O-Dang! datasets using HurtLex.}
\label{tab:hurtlex-datasets}
\end{table*}

In this way, we have characterized the datasets with HurtLex using a straightforward approach.
For each document, the words that are included in each HurtLex category and in the document are counted.
This outputs a count for each HurtLex category that is related to a document.
To aggregate these counts on a dataset, we average over all documents.

Table~\ref{tab:hurtlex-datasets} shows the result of the described lexical analysis.
Such characterization profiles the use of selected HurtLex categories across all datasets.
One of the most interesting categories in these datasets, due to its prevalence, is CDS (derogatory words).
It can be seen that it is specially relevant in the hate speech and stance datasets.

Continuing with this, the one of the highest metric for the CDS category is obtained in the HaSpeeDe 2018 dataset.
Interestingly enough, when looking at this same metric aggregated by annotation class, we see a shift.
The HaSeeDe 2018 describes a hate speech binary annotation.
For the negative class, the CDS metric has a value of 0.1165 while for the positive class it reaches 0.1546.
This observation gives further insight into the language of the data.

\section{Conclusion and Future Work} \label{sec:conclusion}
In this paper, we presented O-Dang!, a KG of Italian data sets annotated for Dangerous Speech-related phenomena. The KG includes $62,193$ tweets and $258,704$ annotations both aggregated and un-aggregated. The underlying Semantic Model enables to perform comparative analysis between data sets and phenomena. A first exploratory analysis of offensiveness across corpora has also been provided.

Future work will be devoted to employ this resource to fully inform the systems of abusive language detection, gathering useful pragmatic, semantic and interactional patterns.
Moreover, O-Dang! will be integrated with corpora containing different genres of texts in various languages. Finally, a more robust Entity Linking pipeline will be applied, in order to provide more information about Dangerous Speech targets, that may be used for building more explainable systems for abusive language detection.


\section{Acknowledgements}
The work of M. A. Stranisci was funded by the project ``Be Positive!'' (under the 2019 ``Google.org Impact Challenge on Safety'' call).
The work of S. Frenda, V. Patti and C. Bosco was supported by the European project ``{STERHEOTYPES - STudying European Racial Hoaxes and sterEOTYPES}'' funded by the Compagnia di San Paolo and VolksWagen Stiftung under the ``Challenges for Europe'' call for Project (CUP: B99C20000640007). 
The work of A. T. Cignarella and V. Basile was supported by the project "Toxic Language Understanding in Online Communication - BREAKhateDOWN" funded by Compagnia di San Paolo (ex-post 2020). 
The work of Oscar Araque has been funded by the European Union's Horizon 2020 project Participation (grant agreement no. 962547) and the help of the ``Programa Propio'' from ``Universidad Politécnica de Madrid''.









\section{Bibliographical References}\label{reference}

\bibliographystyle{lrec}
\bibliography{bibliografia.bib}

\section{Language Resource References}
\label{lr:ref}
\bibliographystylelanguageresource{lrec}
\bibliographylanguageresource{languageresource.bib}

\end{document}